\pdfoutput=1

\documentclass[11pt]{article}

\usepackage[]{ACL2023}

\usepackage{times}
\usepackage{latexsym}
\usepackage{booktabs}
\usepackage{algpseudocode}
\usepackage{algorithm}

\usepackage{amssymb}
\usepackage{amsthm}
\usepackage{bbm}
\usepackage{booktabs}
\usepackage{marginnote}
\usepackage{subfig}
\usepackage{multirow}
\usepackage{amssymb}
\usepackage{amsthm}
\usepackage{amsmath}
\usepackage{xspace}
\usepackage{thmtools,thm-restate}

\usepackage[T1]{fontenc}

\usepackage[utf8]{inputenc}

\usepackage{microtype}
\usepackage{graphicx}
\usepackage{subfig}

\usepackage{inconsolata}

\usepackage{bm}

\newcommand{\compt}{\ensuremath{t_{\text{score}}}}
\newcommand{\compp}{\ensuremath{t_{\text{parseval}}}}
\newcommand{\phen}{structural grokking\xspace}

\newcommand{\qf}{Question-Formation}
\newcommand{\ti}{Tense-Inflection}

\newcommand{\spn}{\ensuremath{p}}

\newcommand{\sci}{\ensuremath{\mathrm{SCI}}}

\newcommand{\cvec}[2]{\ensuremath{\vv^{#1}_{#2}}}

\newcommand{\treeo}[1]{\ensuremath{\widehat{T}_{\text{proj}}(#1)}}

\newcommand{\senti}[1]{\ensuremath{S^{(i)}}}

\DeclareMathOperator*{\argmin}{arg\,min}

\def\vv{{\bm{v}}}

\title{Grokking of Hierarchical Structure in Vanilla Transformers}

\author{
\textbf{Shikhar Murty}\textsuperscript{$\dagger$}
\hspace{.1cm}\textbf{Pratyusha Sharma}\textsuperscript{$\ddagger$} \hspace{.1cm} \textbf{Jacob Andreas}\textsuperscript{$\ddagger$} \hspace{.1cm} \textbf{Christopher D. Manning}\textsuperscript{$\dagger$} \\
\textsuperscript{$\dagger$}Computer Science Department, Stanford University\quad
  \textsuperscript{$\ddagger$}MIT CSAIL\\
  \texttt{\{smurty, manning\}@cs.stanford.edu, \{pratyusha, jda\}@mit.edu} 
}

\begin{document}

\maketitle

\begin{abstract}
For humans, language production and comprehension is sensitive to the hierarchical structure of sentences. In natural language processing, past work has questioned how effectively neural sequence models like transformers capture this hierarchical structure when generalizing to structurally novel inputs. We show that transformer language models can learn to generalize hierarchically after training for extremely long periods---far beyond the point when in-domain accuracy has saturated. We call this phenomenon \emph{\phen}. 
On multiple datasets, \phen{} exhibits inverted U-shaped scaling in model depth: intermediate-depth models generalize better than both very deep and very shallow transformers. When analyzing the relationship between model-internal properties and grokking, we find that optimal depth for grokking can be identified using the tree-structuredness metric of \citet{murty2023projections}. Overall, our work provides strong evidence that, with extended training, vanilla transformers discover and use hierarchical structure.
\end{abstract}

\section{Introduction}

\begin{figure}[t]
    \centering
    \includegraphics[width=\linewidth]{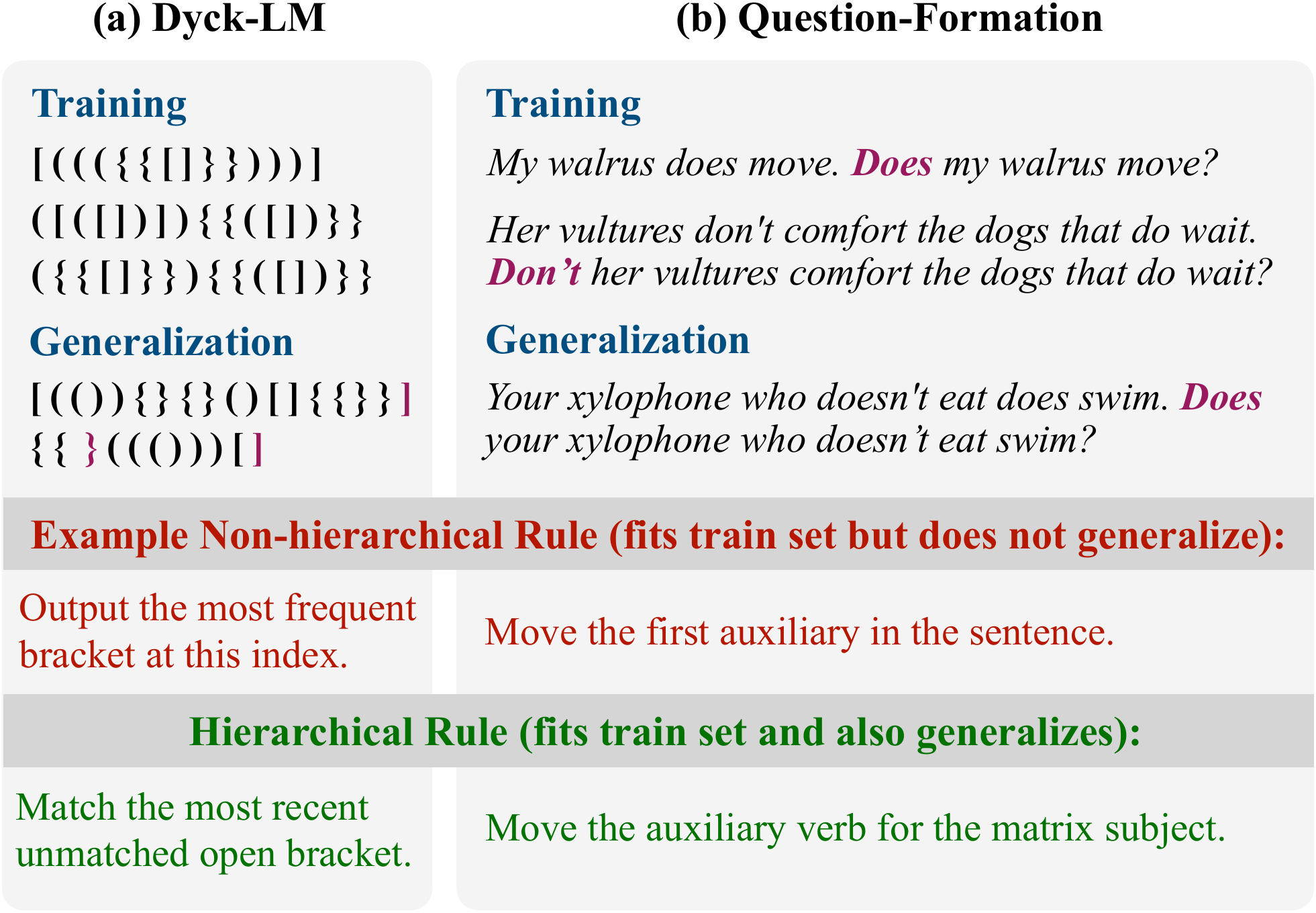}
    \caption{Examples from language modeling datasets we use to assess hierarchical generalization in vanilla transformers. These datasets are constructed so that both a non-hierarchical as well as a hierarchical rule can perfectly fit the training set, but only the hierarchical rule generalizes to structurally novel inputs.}
    \label{fig:task}
\end{figure}

Although human language is produced as a linear sequence, it is hierarchically organized. Smaller units compose to form larger constituents. The ability to infer this hierarchical structure underlies our ability to produce and understand new sentences \citep{chomsky1965, crain1987structure}. In this paper, we investigate whether standard neural transformer models \cite{vaswani2017transformer} can also generalize hierarchically when trained on language processing tasks (Fig~\ref{fig:task}). 
Our main finding is that hierarchical generalization in transformers does occur, but very slowly: performance on structurally novel sentences increases gradually, long after performance on sentences from the training distribution has plateaued.
We term this phenomenon \emph{\phen{}}, by analogy to existing findings on simple classification tasks \citep{power2022grokking}.

On two datasets, we show that \phen{} exhibits inverted U-shaped scaling behavior as a function of model depth: hierarchical generalization improves, then declines, as we train deeper models. Prior work suggests that a number of model-internal properties might track the emergence of hierarchical structure in transformers, including weight norms \citep{merrill-etal-2021-effects, liu2022omnigrok, power2022grokking}, attention sparsity \citep{merrill-etal-2021-effects}, and functional tree-structuredness \citep{murty2023projections}. We find that functional tree-structuredness is uniquely able to predict \phen---while weight norms and attention sparsity increase monotonically in model depth,
tree-structuredness is highest for models of the optimal depth for \phen{}. 

Our results challenge findings from prior work  \citep{mueller-etal-2022-coloring, petty2021transformers} claiming that ordinary transformers completely fail on the tests of hierarchical generalization that we study. 
We attribute these failures to early stopping based on in-domain validation performance, which significantly underestimates hierarchical generalization due to \phen{}. On the datasets where this prior work reports generalization accuracies below 20\%, 
 \emph{simply by training for longer},
mean accuracy across random seeds reaches 80\%, and several seeds achieve near-perfect generalization performance. Past findings are also partially explained by U-shaped scaling: this work uses models that are too shallow \citep{mueller-etal-2022-coloring, petty2021transformers} or too deep \citep{mueller-etal-2022-coloring}. 
Our results align with past findings on the role of extended training in other language processing problems \citep{csordas-etal-2021-devil,hoffmann2022training}.

\section{Background}

\paragraph{Transformers} Given a sequence of tokens $w_{\leq{i}} = w_1, w_2, \ldots, w_{i}$, where each token is drawn from a fixed vocabulary $V$, an $L$-layer transformer language model (LM) $f_{\theta}^{L}$ outputs a distribution over the next token $w_{i+1} \in V$, $f_\theta^{L}(w_{\leq{i}}) \in \mathbb{R}^{|V|}$. A key part of the architecture is a sequence of $L$ \emph{self-attention} layers, where layer $p$ computes contextual vectors of token $k$ as a non-linear parametric function of a convex combination of contextual vectors of tokens $w_{\leq{k}}$ from the previous layer, where coefficients $\mathbf{a}^{p}_{k} \in \mathbb{R}^{k}$ are known as the \emph{attention distribution}. The LM weights are learned by maximizing the log probability of the correct continuation $w_{k+1}$, given prefix $w_{\leq{k}}$.

\paragraph{Hierarchical structure in transformers}
While unsupervised pre-training of transformers 
has led to state-of-the-art transfer learning results across NLP, the architecture itself has been claimed to lack human-like inductive biases toward hierarchical structure \citep{tran2018importance, hahn2020theoretical, petty2021transformers, mueller-etal-2022-coloring}. We revisit these claims in this work. 

To understand whether a given model has a bias for acquiring hierarchical structure, we follow \citet{mccoy2020syntax} and evaluate generalization in models trained on ambiguous tasks in which training data is consistent with both a ``hierarchical rule'' as well as a ``non-hierarchical rule'' (Fig~\ref{fig:task}). To test if the hierarchical rule has been acquired, we test generalization on a separate out-of-distribution test set, constructed such that only learners that have acquired the hierarchical rule are successful.

\paragraph{Grokking} \citet{power2022grokking} identify the phenomenon of \emph{grokking} on small algorithmic datasets where they find that test performance improves long after training performance has saturated. We hypothesize a similar \emph{\phen{}}, where the model groks hierarchical structure long after in-domain validation performance has saturated, and consequently, hierarchical generalization can continue to improve with extended training.

\begin{figure*}[!ht]
\centering
    \includegraphics[width=\linewidth]{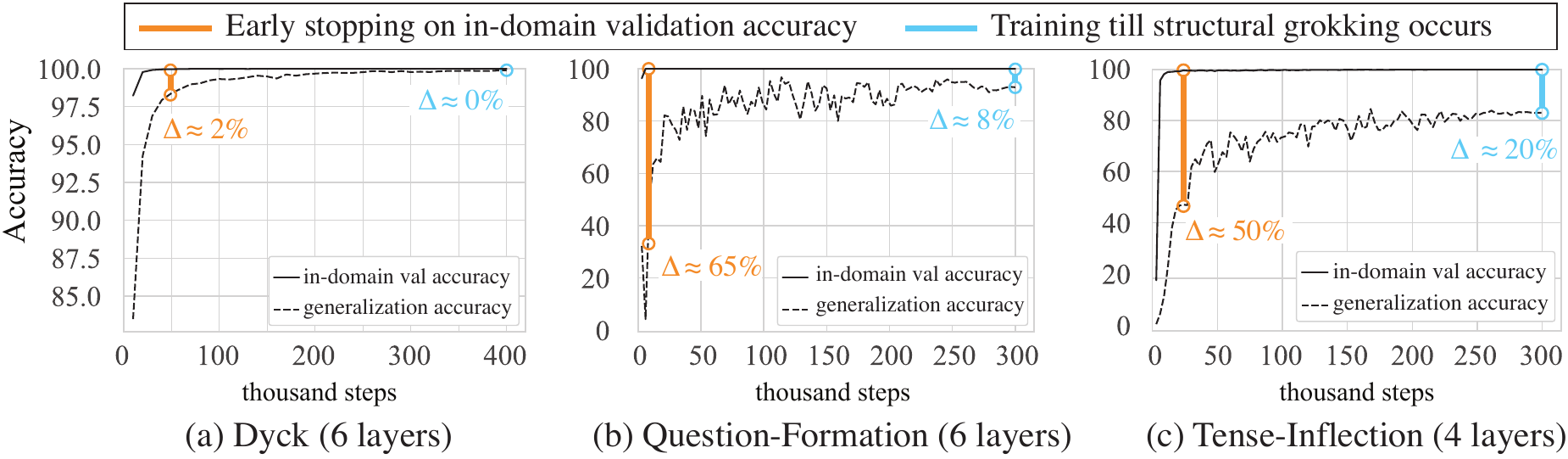}
\caption{Average accuracy across 10 random seeds on the in-domain val set (solid) and generalization set (dashed) for all datasets. Generalization performance improves even after in-domain accuracies have saturated, showing \emph{\phen{}}. We highlight with orange and blue lines the gap between in-domain and generalization accuracies at the point of early stopping based on the in-domain val set performance vs.\ at the end of training, noting that prior work stops training at the orange line. Stopping training prior to \phen{} can result in a vast underestimation of generalization performance.}
\label{fig:grokking_main_result}
\end{figure*}

\begin{figure*}
\centering
\includegraphics[width=\linewidth]{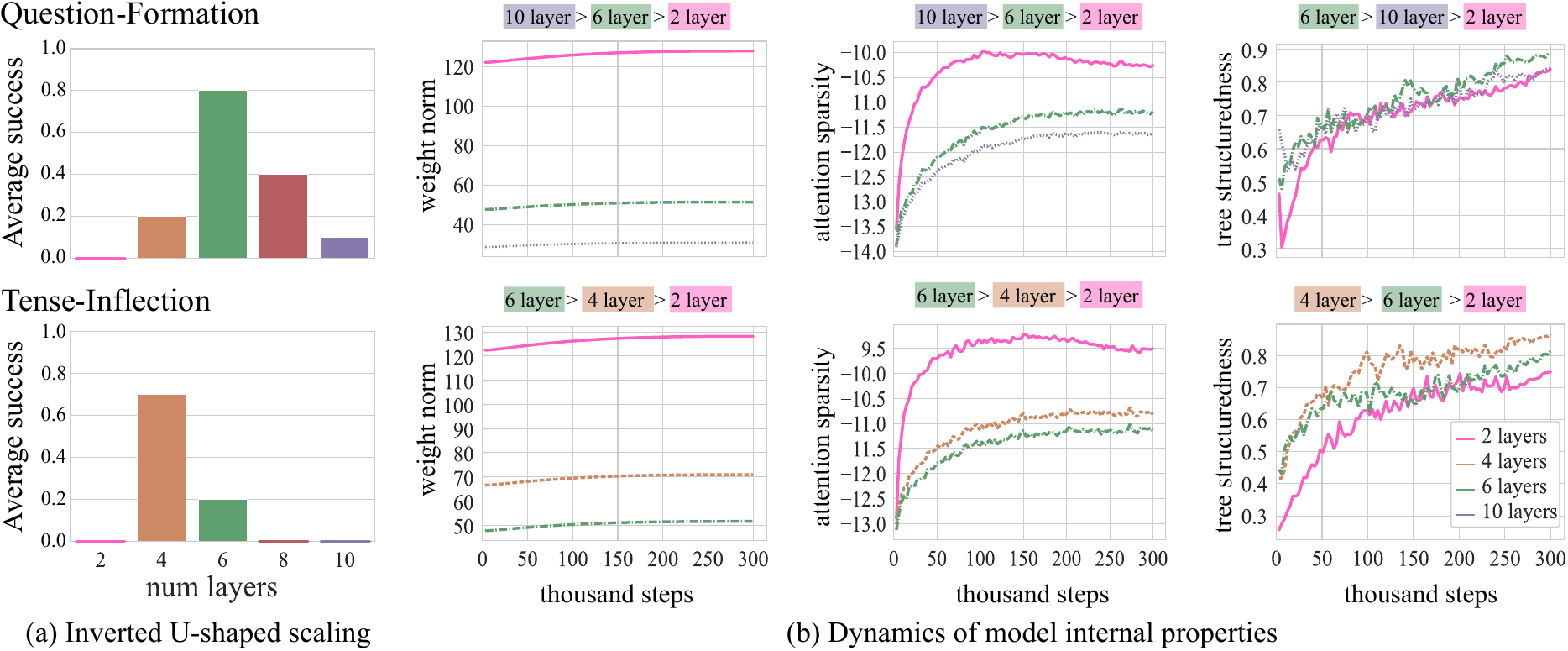}

\caption{(a) Inverted U-shaped laws for grokking: On \qf{} (top) and \ti{} (bottom), we find that both very small and very deep models either fail to exhibit \phen or do so infrequently, compared to an in-between optimal model depth. (b) While weight norms and attention sparsity increase for all models and do not differentiate between different sizes, tree-structuredness is highest for the optimal model depth.}
\label{fig:u-shaped-and-model-internal}
\end{figure*}

\section{Experiments}

\paragraph{Datasets} Since our goal is to understand hierarchical generalization in transformers, we use two datasets from \citep{mccoy2020syntax} and additionally evaluate on a simple bracket-tracking task. For \emph{Dyck}, models are trained to predict next tokens in strings drawn from $\mathrm{Dyck}_{20, 10}$, the language of well-nested brackets with $20$ types and max nesting depth of $10$. We evaluate generalization to structurally unobserved strings in $\mathrm{Dyck}_{20, 10}$ (see Fig~\ref{fig:task} for examples and Appendix~\ref{sec:dataset_details} for details). For the \citet{mccoy2020syntax} datasets, in \emph{\qf}, models must convert English sentences into questions and, in \emph{\ti}, models must map from sentences and tense markers to appropriately re-inflected sentences. We evaluate generalization on the out-of-distribution test set from \citet{mccoy2020syntax}.

\paragraph{Model} We train transformer LMs with \{2, 4, 6, 8, 10\} layers (see Appendix~\ref{sec:model_details} for more details). For each depth, we train models with 10  random seeds for 300k (400k for Dyck) steps. Given the input sentence (or prefix in the case of Dyck) we decode greedily from the model at test time. For Dyck, we report the accuracy of generating the correct closing bracket type by ranking among closing brackets, given an input prefix from the language. As done in prior work \citep{mccoy2020syntax, petty2021transformers, mueller-etal-2022-coloring}, for \qf{}, we report first word accuracy of the decoded question, and for \ti{}, we report the fraction of test inputs for which the target verb is correctly inflected. 

\subsection{Main Results}

\paragraph{Transformers exhibit \phen} We first present results obtained with the best model depth on all datasets in Fig~\ref{fig:grokking_main_result}. We find clear evidence of \phen{}: Across datasets, generalization improves many training steps after in-distribution accuracy has saturated, sometimes approaching perfect accuracy.

\paragraph{Early stopping considered harmful} Next, we compare generalization accuracy obtained by early stopping on in-domain validation accuracy (as done in \citet{petty2021transformers, mueller-etal-2022-coloring}) to longer training runs (Fig~\ref{fig:grokking_main_result}). Early stopping leads to vastly underestimating generalization. For instance, average generalization goes up from <40\%, <50\% to  <90\%, <80\% on \qf{} and \ti{}, respectively. 

\paragraph{Inverted U-shaped scaling} On \qf{} and \ti{}, we train models of increasing depths from 2 to 10 layers. For each depth, we report the fraction of seeds (out of 10) where generalization accuracy eventually crosses 80\%, in Fig~\ref{fig:u-shaped-and-model-internal}a. We find an inverted U-shaped scaling behavior---very shallow and very deep models are unsuccessful, while most seeds generalize in models of intermediate depth. This may also explain why prior work that either used very shallow models (1--3-layer transformers in \citet{petty2021transformers, mueller-etal-2022-coloring}) or very deep models (12-layer transformers in \citet{mueller-etal-2022-coloring}) failed to generalize well.

\section{Analysis}
\label{sec:model_props}

Given that \phen occurs only in a subset of model architectures, can we identify when it has happened (or predict when it will occur)?
Several model-internal properties have been claimed to relate to either grokking or emergent hierarchical structure in transformers.

\paragraph{Weight Norms} Recent work \citep{power2022grokking, liu2022omnigrok} identifies the $L_2$ norm of parameter weights as an important quantity for grokking. For instance, \citet{power2022grokking} find weight decay to improve grokking speed and \citet{liu2022omnigrok} identify a ``goldilocks zone'' in weight norm space where grokking occurs. More generally, norm growth over the course of training has been studied as a key factor in neural network generalization \citep{soudry2018implicit}. 
\paragraph{Attention Sparsity} 
\citet{merrill-etal-2021-effects} prove that norm growth in transformers leads to attention saturation, an important property for emergent linguistic structure \citep{merrill-etal-2022-saturated}. As a proxy for attention sparsity of $f_\theta^{L}$, we compute the negative mean entropy of all distributions $\{\mathbf{a}^{p}_{k}\}$. 

\paragraph{Tree-structuredness}  \citet{mccoy2020syntax} show that tree-structured encoders such as \citet{tai-etal-2015-improved} show near perfect hierarchical generalization. While transformers are relatively unconstrained, recent evidence suggests that, when trained on language data, they implictly implement (approximately) tree-structured computations. In particular, the \emph{tree projection} method of \citet{murty2023projections} precisely characterizes the extent to which a transformer's internal computation on an input can be approximated with a tree-structured neural encoding, providing a tree-structuredness score (\compt{}) for any transformer, and a binary tree that best approximates its computation on an input string (see Appendix~\ref{sec:func_ts} for details). To evaluate whether these trees correspond to human notions of syntax, we additionally compare recovered trees to gold-standard ones \citep[\compp{},][]{black1991procedure}. 

\subsection{Results}

We characterize the \emph{dynamics} of weight norms (normalized by number of layers to compare different model depths), attention sparsity, and tree-structuredness, by computing these quantities every 3k gradient updates for \qf{} and \ti{}. For data-dependent properties such as attention sparsity and tree-structuredness, we sample 10k examples from the training data. We plot these quantities for the smallest model, the largest model for which at least one run shows successful grokking, and for the optimal model depth, in Fig~\ref{fig:u-shaped-and-model-internal}b.

\paragraph{Optimal models are most tree-structured} Weight norms and attention sparsity grow for all model settings in both datasets. However, these properties by themselves are unable to predict that both shallow and deep models fail---shallow models learn the sparsest solutions as well as solutions with largest weight norms, but never generalize hierarchically. As noted by \citet{murty2023projections}, $\compt{}$ improves over time for all models, indicating increased tree-structuredness over time. For both datasets, the ``optimal'' model learns the most tree-structured solution compared to both deep and shallow models. \citet{liu2022omnigrok} note that, on algorithmic tasks, grokking ``coincides with the emergence of structure in embeddings''. Similarly, for language tasks, we find that structural grokking coincides with the emergence of tree structured internal computations.

\begin{figure}
\centering
\includegraphics{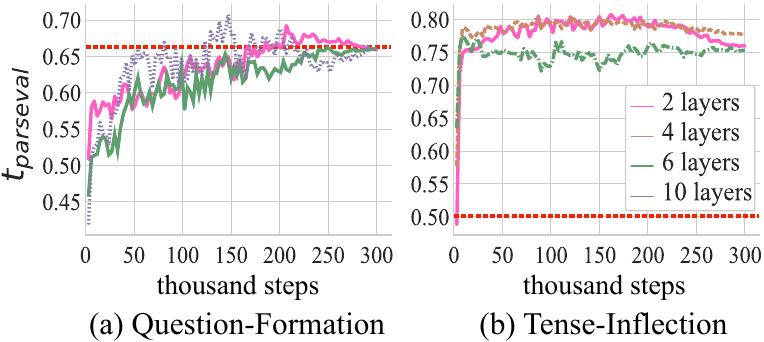}
\caption{While some models fail to generalize hierarchically, all models are effective at learning computations whose closest tree structures progressively evolve towards ground truth syntax, matching or outperforming a right branching baseline (in dashed red).}
\label{fig:structure_induction}
\end{figure}

\paragraph{Transformers are surprisingly effective at structure induction} From the dynamics of $\compp{}$ in Fig~\ref{fig:structure_induction}, we note that all models, \emph{regardless of whether they generalize or not}, learn structures that are close to ground truth syntax, sometimes outperforming a right-branching baseline. \citet{mccoy2020syntax} note that tree-structured encoders only generalize when structured according to correct parse trees. Here, we find that all transformers learn correct tree structures, but only the ones that are the most tree-structured generalize best.

\section{Conclusion}

This work shows that transformers are capable of exhibiting \emph{structure-sensitive} ``hierarchical generalization'' via a grokking mechanism. Their overall learning behavior gradually shifts from memorization (high in-domain accuracy, poor out-of-domain accuracy) to generalization (high in-domain and out-of-domain accuracy). While we show such behavior on relatively small datasets with small models, we believe these results may have broader implications, as training for longer has been shown to help even for web-scale language modeling \citep{hoffmann2022training} and on compositional generalization tasks \citep{csordas-etal-2021-devil}. Structural grokking happens most often at ``medium-sized'' model depths, and both very shallow and very deep models fail to exhibit it. While properties previously connected with linguistic generalization in transformers such as weight norms and attention sparsity do not differentiate good architectures from bad ones, functional tree-structuredness of the transformer can well predict the optimal model depth. While there are clear limitations to the transformer architecture (such as the inability to implement unbounded recursion), our results show that it may have stronger inductive biases than previously believed: With sufficient training, transformers can represent hierarchical sentence structure and use this structure to generalize correctly.

\section{Reproducibility}

All code and data for these experiments is available at \href{https://github.com/MurtyShikhar/structural-grokking.git}{\url{https://github.com/MurtyShikhar/structural-grokking.git}}.

\section{Acknowledgements}

SM was funded by a gift from Apple Inc. CM is a fellow in the CIFAR Learning in Machines and Brains program. We thank John Hewitt, Belinda Li, Rishi Bommasani and members of the Stanford NLP group for feedback on the paper.

\section*{Limitations}
Our work has the following limitations. First, we only evaluate generalization on datasets based on English language. Second, we show \phen{} on three datasets, and while we believe this to be a general phenomenon, we leave investigating similar behavior on other datasets for future work. Next, we also do not study the effect of training data size on structural grokking, and do not investigate whether transformers learn to grok hierarchical structure in low data regimes. Finally, all datasets here are based on context-free grammars, either similar to or taken directly from prior work, and we believe constructing similar generalization benchmarks on real language data is a good avenue for future work.

\bibliography{anthology,custom}
\bibliographystyle{acl_natbib}

\clearpage
\appendix

\section{Dataset Details}
\label{sec:dataset_details}

\begin{table}
\centering
\tiny
\begin{tabular}{@{}lccc@{}} \toprule
\textbf{Dataset}       & Train & In-Domain Val & Generalization \\ \midrule
Dyck  & 200000 & 20000 & 20000 \\ 
\qf{}  & 100000 & 1000 & 10000 \\
\ti{} & 100000 & 1000 & 10000  \\\bottomrule
\end{tabular}
\caption{Statistics for all datasets used in this work.}
\label{tab:stats}
\end{table}

All statistics are in Table~\ref{tab:stats}. For \qf{} and \ti{}, we use splits as given in \citet{mccoy2020syntax} with no additional preprocessing. We give details of Dyck below.

\paragraph{Dyck Details} We construct our Dyck dataset by sampling 200k strings from $\mathrm{Dyck}_{20,10}$, the language of well-nested brackets with 20 different bracket types and nesting depth atmost 10. 
For each string, we define its structure as a binary vector of 0s and 1s. For instance the structure of ``(({[]}))'' is ``11110000''. To construct a generalization set, we sample strings with unobserved structures i.e. strings whose 0-1 structure does not match the structure of any of the training strings. Since the objective at test time is to measure closing bracket accuracy, we only rank model probability among all closing brackets, and only evaluate on prefixes where the opening bracket is atleast 10 tokens away from its corresponding closing bracket.

\section{Model Details}
\label{sec:model_details}
We use a transformer language model with the following hyperparameters:
\begin{itemize}
    \item Number of attention heads = 4
    \item Hidden dimensionality = 512 
    \item Tied input and output matrices as done in \citet{press2017using}
\end{itemize}

Next, we use the following hyperpameters for optimization:
\begin{itemize}
    \item AdamW ($\beta_1$: 0.9, $\beta_2$: 0.999, $\epsilon$: 1e-7),  with learning rates in \{1e-4, 5e-5, 1e-5\}, noting that 1e-4 works best for all experiments. We use a linear warmup scheduler warming up from 0 to the final learning rate over 10k gradient steps.
    \item We clip gradients to have a max $L_2$ norm of 10.
    \item We use a batch size of $8$.
\end{itemize}

\section{Functional Tree-Structuredness}
\label{sec:func_ts}

Tree Projections (TP; \citet{murty2023projections}) measure how well computations performed by a given transformer $f$ can be approximated with tree-structured encoders. To do this, TP solves the following optimization problem:
\begin{align}
\label{eq:1}
    \phi_{\text{proj}}, T_{\text{proj}} \triangleq \argmin_{\phi,T} \mathcal{L}(f, g_\phi, T),
\end{align} 

\noindent where $g_\phi$ is the class of tree structured encoders that processes sentence $S$ according to bottom-up trees $T(S)$, and $\mathcal{L}$ is a distance function between vector outputs of $f$ and $g_\phi$ on spans from the binary tree $T$. TP minimizes Equation~\ref{eq:1} approximately, and recovers an approximate $\widehat{T}_{\text{proj}}$. The tree score over a dataset $\mathcal{D}$ is defined as
\begin{align}
\label{eq:compt_def}
\compt{} \triangleq \frac{\sum_{S \in \mathcal{D}} \mathbb{E}_T{\sci{}(S, T)} - \sci{}(S, \treeo{S})}{|\mathcal{D}|},
\end{align}

\noindent where $\sci$ (span contextual invariance) is the distance between contextual and context-free vector representations of all spans $\spn$ in $T$ (for more details, see \citet{murty2023projections}). In particular, $\sci$ score for a sentence $S$ structured according to $T(S)$ is
\begin{align}
    \sci{}(S, T) \triangleq \sum_{s \in T} d(\cvec{S}{\spn},\tilde{\vv}_{\spn})
\end{align}

\noindent for some suitably chosen distance function $d$ (here, cosine similarity). To measure the bracketing F1 score (PARSEVAL; \citet{black1991procedure}) of the induced tree projection of the transformer $\widehat{T}_{\text{proj}}$ against ground truth gold syntax trees, $T_g$, when available, \citet{murty2023projections} define 
\begin{align}
 \compp{} \triangleq \textsc{PARSEVAL}(\widehat{T}_{\text{proj}}, T_g, \mathcal{D}).   
\end{align}

\section{Training Loss Curves}
\begin{figure*}[ht]
\centering
\includegraphics[width=\linewidth]{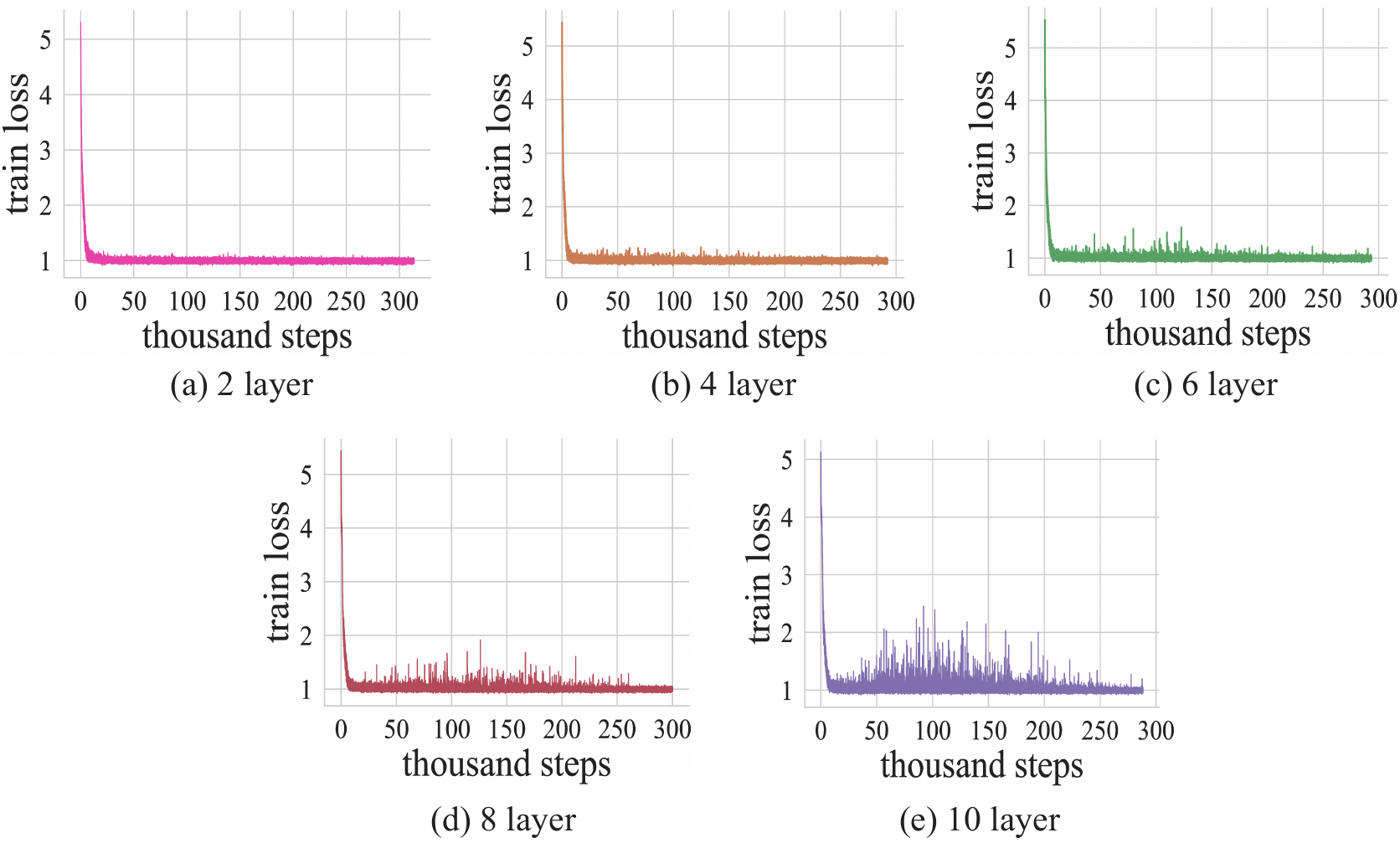}
\caption{We plot average training loss for all model depths on the \qf{} dataset. We note that (1) grokking happens even though training losses fully flat line around 10k gradient steps, and that (2) the inverse U-shaped scaling is not a result of poor optimization of small / large models since all models eventually have the same stable training loss as the optimal model depth.}
\label{fig:training-loss}
\end{figure*}

We explore the hypothesis that syntactic grokking is simply a result of the training loss continuing to decrease, even after in-domain validation performance has saturated in Fig~\ref{fig:training-loss}. We note that training losses generally saturate before in-domain validation performance saturates (also noted in \citet{power2022grokking}). Next, we also find that all models, regardless of whether they grok or not, eventually get to comparable training losses. We conclude that the inverted U-shaped trend is not an artifact of poorly optimized models.

\end{document}